\pdfoutput=1

\documentclass[11pt]{article}

\usepackage[]{ACL2024}

\usepackage{times}
\usepackage{latexsym}
\usepackage{booktabs}
\usepackage{multirow}
\usepackage{enumitem}
\usepackage[frozencache,cachedir=minted-cache]{minted}
\usepackage{changepage}
\usepackage{colortbl}
\definecolor{light-gray}{rgb}{0.9,0.9,0.9}

\usepackage[T1]{fontenc}

\usepackage[utf8]{inputenc}

\usepackage{microtype}

\usepackage{inconsolata}

\usepackage{graphicx}

\title{
\raisebox{-3.0pt}{\includegraphics[scale=0.12]{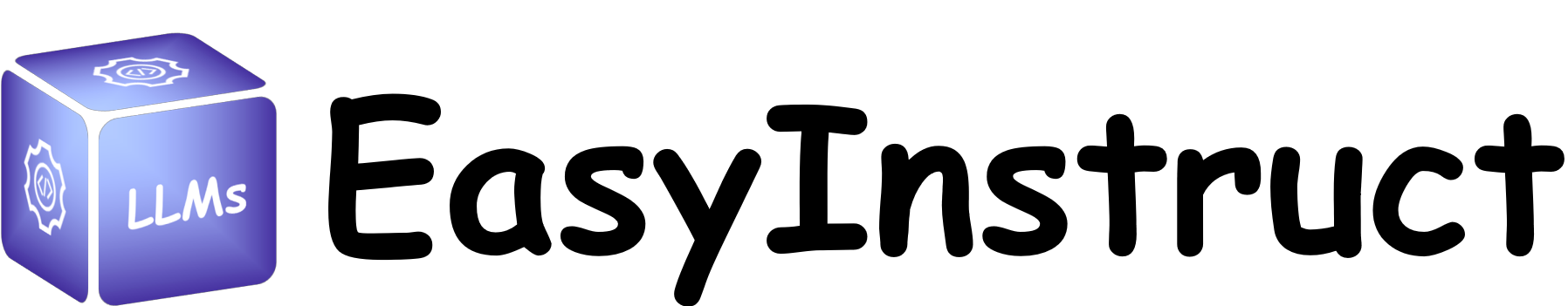}}: An Easy-to-use Instruction Processing Framework \\for Large Language Models
}

\author{
Yixin Ou{$^{\spadesuit}$}, Ningyu Zhang{$^{\spadesuit} \footnotemark[1]$}, Honghao Gui{$^{\spadesuit}$}, Ziwen Xu{$^{\spadesuit}$}, \\
{\bf Shuofei Qiao}{$^{\spadesuit}$},
{\bf Yida Xue}{$^{\spadesuit}$},
{\bf Runnan Fang}{$^{\spadesuit}$},
{\bf Kangwei Liu}{$^{\spadesuit}$},\\
{\bf Lei Li}{$^{\spadesuit}$},
{\bf Zhen Bi}{$^{\spadesuit}$}, 
{\bf Guozhou Zheng}{$^{\spadesuit}$}, 
{\bf Huajun Chen}$^{\spadesuit}\thanks{~~~Corresponding Author.}$\\
 $^\spadesuit$ Zhejiang University \\
  \texttt{\{ouyixin,zhangningyu,huajunsir\}@zju.edu.cn}\\
  \raisebox{-1.2pt}{\includegraphics[scale=0.03]{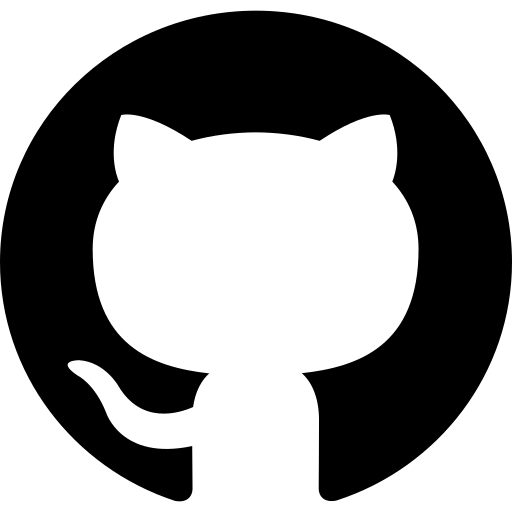}}\,\url{https://zjunlp.github.io/project/EasyInstruct}
  }

\begin{document}
\maketitle

\begin{abstract}

In recent years, instruction tuning has gained increasing attention and emerged as a crucial technique to enhance the capabilities of Large Language Models (LLMs). To construct high-quality instruction datasets, many instruction processing approaches have been proposed, aiming to achieve a delicate balance between data quantity and data quality. Nevertheless, due to inconsistencies that persist among various instruction processing methods, there is no standard open-source instruction processing implementation framework available for the community, which hinders practitioners from further developing and advancing. To facilitate instruction processing research and development, we present \ours\footnote{This is a subprobject of KnowLM (\url{https://github.com/zjunlp/KnowLM}), which facilitates knowledgeable LLM Framework with EasyInstruct, EasyEdit \cite{DBLP:journals/corr/abs-2308-07269,DBLP:journals/corr/abs-2305-13172,DBLP:journals/corr/abs-2401-01286}, EasyDetect etc.}, an easy-to-use instruction processing framework for LLMs, which modularizes instruction generation, selection, and prompting, while also considering their combination and interaction. EasyInstruct is publicly released and actively maintained at \url{https://github.com/zjunlp/EasyInstruct}, along with an online demo app\footnote{\url{https://huggingface.co/spaces/zjunlp/EasyInstruct}} and a demo video\footnote{\url{https://youtu.be/rfQOWYfziFo}} for quick-start, calling for broader research centered on instruction data and synthetic data.
\end{abstract}

\begin{figure*}[htbp]
    \centering
    \resizebox{\linewidth}{!}{
        \includegraphics{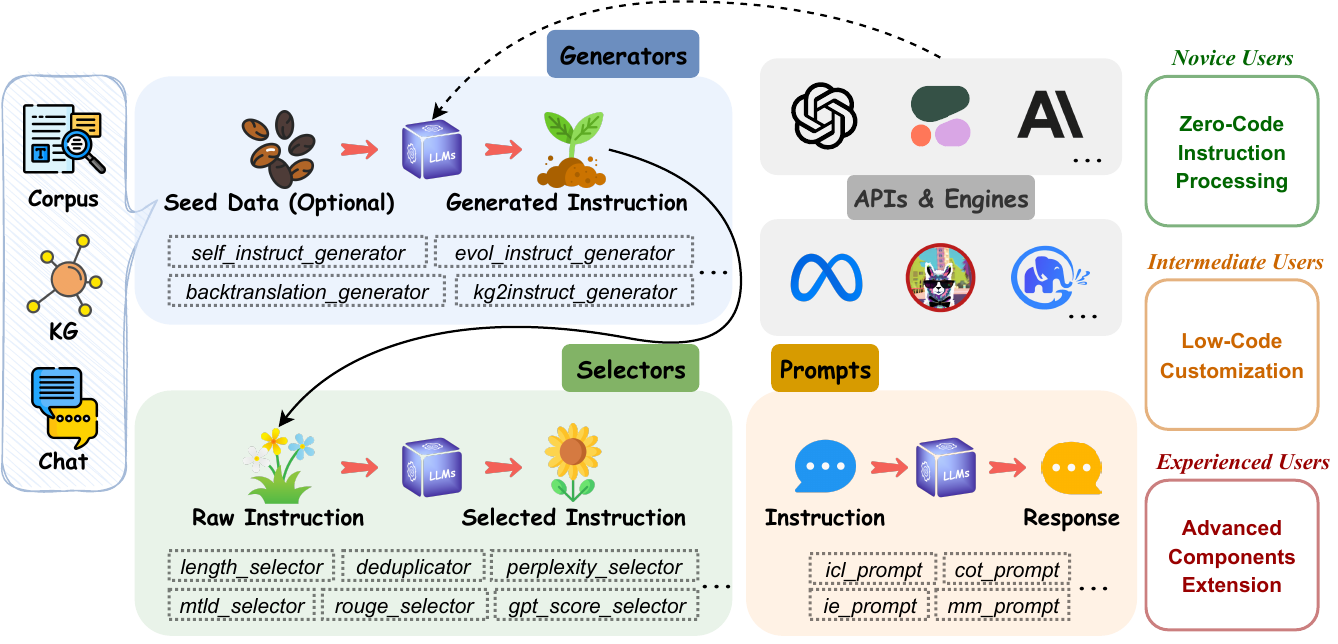}
    }
    \caption{Overview of \raisebox{-1.8pt}{\includegraphics[scale=0.07]{figures/EasyInstruct.png}}. The \texttt{APIs \& Engines} module standardizes the instruction execution process, enabling the execution of instruction prompts on the LLM API services or locally deployed LLMs. The \texttt{Generators} module streamlines the instruction generation process, enabling automated generation of instruction data based on \textbf{chat data, corpus, or Knowledge Graphs}. The \texttt{Selectors} module standardizes the instruction selection process, which enables the extraction of high-quality instruction datasets from raw, unprocessed instruction data. The \texttt{Prompts} module standardizes the instruction prompting process.}
    \vspace{-0.1in}
    \label{fig:overview}
\end{figure*}

\section{Introduction}

Large Language Models (LLMs) have brought about a revolutionary transformation in the field of Natural Language Processing (NLP), leading to substantial improvement in performance across various tasks~\citep{GPT-3, GPT-4, PaLM2, Llama2, LLM-survey, DBLP:conf/www/ChenZXDYTHSC22, reasoning-survey, LKM-survey}.
To optimize the performance of LLMs in specific tasks or domains, it is crucial to adapt their outputs to specific contexts or instructions.
Recent studies~\citep{Flan, InstructGPT, Scaling} have proposed instruction tuning methods for fine-tuning LLMs, which is a prominent research area aimed at optimizing the LLMs' behavior by providing explicit instructions during training, enabling better control and alignment with user preferences and desired outputs.
Instruction dataset construction, which is also referred to as data engineering or management, poses a significant challenge in the process of instruction tuning~\citep{LLM-survey, IT-survey, Align-Survey, DM-survey}.

Substantial efforts have been dedicated to the task of construction instruction data through human annotations~\citep{Super-NaturalInstructions, OpenAssistant}, requiring a significant allocation of resources.
Against this backdrop, LLMs are utilized to synthesize large-scale instruction data automatically
~\citep{Self-Instruct, WizardLM, Backtranslation}.
These methods could scale up the size of instruction-following data, but they still inevitably suffer limited diversity and complexity, resulting in an unbalanced distribution and poor quality of instruction data.
Recent studies~\citep{LIMA, AlpaGasus, WizardLM} have unveiled a seminal revelation, indicating that even a small quantity of high-quality instruction data has the potential to yield robust performance.
In general, instruction processing is an important process requiring careful attention to detail and rigorous quality assurance procedures to construct a high-quality instruction dataset for LLMs.

Unfortunately, the availability of open-source tools for instruction processing remains limited, especially in comparison to many open-source projects on models and training infrastructures~\citep{Llama, Llama2, Alpaca, BLOOM, vicuna2023, GLM-130B}.
Existing projects are often highly-customized to their own needs, lacking a systematized and modular processing ability to address diverse processing pipelines for LLMs.
For instance, the Alpaca~\citep{Alpaca} dataset targets the augmentation of diversity for LLaMA tuning, whereas AlpaGasus~\citep{AlpaGasus} focuses on filtering out low-quality instances from Alpaca.
Thorough development of instruction processing systems for the ever-evolving and emerging requirements of LLM remains unexplored, particularly in light of the quick expansion of inventive LLM applications spanning various fields.

To address this issue, we develop EasyInstruct as depicted in Figure~\ref{fig:overview}, an easy-to-use instruction processing framework for LLMs.
Given some existing \textbf{chat data, corpus, or Knowledge Graphs}, EasyInstruct can handle instruction generation, selection, and prompting processes, while also considering their combination and interaction. 
These consistencies facilitate further development and comparisons of various methods, thus promoting the advancement of better instruction processing work.
We further conduct experiments with EasyInstruct to validate its effectiveness in instruction processing.
Currently, EasyInstuct is open-sourced on GitHub and has already received over 300 stars.
We are committed to the long-term maintenance of EasyInstruct, providing continuous support for new features to ensure its effectiveness as a framework for instruction processing and synthetic data generation~\citep{synthetic_survey}.

\section{Background}

LLMs typically undergo two stages of training: pre-training and fine-tuning~\citep{LLM-survey}.
Despite the fact that large-scale pretraining is the key of the model's proficiency in generating natural language responses, these pre-trained models can still struggle with comprehending human instructions accurately.
To bridge the gap between the training objectives and human objectives, instruction tuning is introduced as a potent strategy to amplify the controllability and capabilities and  of LLMs in interpreting and responding to instructions~\citep{Flan, InstructGPT, Scaling, Self-Instruct, IT-survey, IL-survey}.
Concretely, instruction tuning involves the method of refining pre-trained LLMs through supervised learning, utilizing examples structured as \textsc{(Instruction, Input, Output)}. 
In this format, \textsc{Instruction} represents the human-given directive that outlines the task, \textsc{Input} optionally offers additional context, and \textsc{Output} signifies the expected outcome in alignment with the \textsc{Instruction} and any given \textsc{Input}.

Despite the effectiveness of instruction tuning, constructing high-quality large-scale instructions which effectively encompass the target behaviors remains a non-trivial challenge in this realm.
Existing instruction datasets are often limited in terms of diversity, quantity, and creativity, which underscores the significance of instruction processing.
One typical method for constructing instruction datasets is data integration.
In this method, instructional datasets are constructed by merging existing annotated datasets with descriptions of tasks in natural language \citep{Flan-Collection, P3, gpt4all}.
Another prevalent method for constructing instruction datasets is automated generation. 
To alleviate the need for extensive human annotation or manual data gathering, automated methods have been proposed to generate large volumes of instructional data through the use of LLMs.
Instructions can be sourced from chat data~\citep{vicuna2023} or expanded on a small set of seed instructions using LLMs~\citep{Self-Instruct, WizardLM, Backtranslation}. 
Subsequently, the collected instructions are fed into LLMs to generate corresponding inputs and outputs.
In EasyInstruct, our primary focus lies on automated approaches for instruction generation due to their high efficiency and scalability.



Another promising research direction of instruction processing is the selection of high-quality instruction.
Recently, numerous studies~\citep{LIMA, AlpaGasus, WizardLM, deita} have investigated the issue of the scale of the instruction dataset for fine-tuning and have indicated that merely increasing the number of instructions may not necessarily result in enhancements.
Instead, a modest volume of high-quality instruction data can  influence the fine-tuning of LLMs, yielding solid performance.
Thus, optimizing the instruction dataset and enhancing its quality play a critical role in fine-tuning LLMs effectively.

From a practical implementation point of view, instruction processing is actually complex and requires meticulous consideration. 
In this paper, we present \ours, an easy-to-use framework to effectively and efficiently implement instruction processing approaches including instruction generation, selection, and prompting. 
Through this framework, EasyInstruct can help users to quickly comprehend and apply the existing instruction processing methods implemented in the package.

\section{Design and Implementation}

As illustrated in Figure~\ref{fig:overview}, EasyInstruct provides a complete instruction processing procedure built on PyTorch and Huggingface.
In this section, we first introduce the design principles, and then detail the implementation of the major modules.

\subsection{Design Principles}

The framework is designed to cater to users with varying levels of expertise, providing a user-friendly experience ranging from code-free execution to low-code customization and advanced components extension options:

\vspace{-0.05in}
\paragraph{Zero-Code Instruction Processing.} 
Novice users, who do not require coding knowledge, can leverage pre-defined configuration files and shell scripts to accomplish code-free instruction processing. 
By running these scripts, they can complete instruction processing tasks without the need for coding skills.
Example configuration files and shell scripts are shown in Appendix~\ref{app:shell_script}.

\vspace{-0.05in}
\paragraph{Low-Code Customization.} 
Intermediate users have the option to customize various process inputs and outputs using a low-code approach. 
This allows them to have more control over the different stages within the framework. A running example is shown in Figure~\ref{fig:code_example}.

\vspace{-0.05in}
\paragraph{Advanced Components Extension.} 
Experienced users can easily extend our components based on their specific scenarios and requirements. 
To customize their classes, users can inherit the base classes of modules and override the necessary methods as per their requirements.
This flexibility enables them to implement their functional components, tailored to their unique needs.

    
    

\subsection{APIs \& Engines}

The \texttt{APIs} modules integrate with mainstream LLMs, including API services provided by companies such as OpenAI\footnote{\url{https://platform.openai.com/docs}}, Anthropic\footnote{\url{https://docs.anthropic.com/claude/docs}}, and Cohere\footnote{\url{https://docs.cohere.com/docs}}.
This integration facilitates the seamless invocation of various relevant steps within the framework.
We list a range of API service providers and their corresponding LLM products that are currently available in EasyInstruct in Appendix~\ref{app:api}.
The \texttt{Engines} module standardizes the instruction execution process, which enables the execution of instruction prompts on several open-source LLMs such as LLaMA~\citep{Llama, Llama2} and ChatGLM~\citep{GLM, GLM-130B}.



\begin{table*}[htbp]
    \renewcommand\arraystretch{1.2}
    \centering
    \resizebox{\linewidth}{!}{
    \begin{tabular}{c|ccp{12.5cm}}
        \toprule
        \textbf{Modules} & \textbf{Methods} & \textbf{Seed} & \textbf{Description}\\
        \midrule
        \multirow{8}{*}{\texttt{Generators}} & \multirow{2}{*}{Self-Instruct} & \multirow{2}{*}{Chat} & The method that randomly samples a few instructions as demonstrations and generates more instructions and input-output pairs using LLM~\citep{Self-Instruct}. \\ 
        & \multirow{2}{*}{Evol-Instruct} & \multirow{2}{*}{Chat} & The method that incrementally upgrades an initial set of instructions into more complex instructions by prompting an LLM with specific prompts~\citep{WizardLM}. \\ \cline{2-4}
        & \multirow{2}{*}{Backtranslation} & \multirow{2}{*}{Corpus} & The method that creates a training instance by predicting an instruction that would be correctly answered by a paragraph in the corpus~\citep{Backtranslation}. \\ \cline{2-4}
        & \multirow{2}{*}{KG2Instruct} & \multirow{2}{*}{KG} & The method that generates Information Extraction (IE) instruction datasets incorporating existing Knowledge Graphs~\citep{InstructIE}. \\ 
        \midrule 
        \textbf{Modules} & \textbf{Metrics} & \textbf{Type} & \textbf{Description}\\
        \midrule
        \multirow{9}{*}{\texttt{Selectors}} & Deduplication & Statistics-based & Repetitive input and output of instances. \\ 
        & Length & Statistics-based & The bounded length of every pair of instruction and output.\\ 
        
        & \multirow{2}{*}{MTLD} & \multirow{2}{*}{Statistics-based} & A metric for assessing the lexical diversity in text, defined as the average length of word sequences that sustain a minimum threshold TTR score~\citep{MTLD}.\\ \cline{2-4}

        & ROUGE & N-gram-based & Recall-oriented understudy for gisting evaluation~\citep{ROUGE}.\\ 
        \cline{2-4}

       & \multirow{2}{*}{CIRS} & \multirow{2}{*}{Structure-based} & The score using the abstract syntax tree to encode structural and logical attributes, to evaluate the correlation between code and reasoning abilities~\citep{CIRS}.\\
       \cline{2-4}

        & Perplexity & LM-based & The exponentiated average negative log-likelihood of text.\\ 
        
        & \multirow{2}{*}{GPT Score} & \multirow{2}{*}{LM-based} & The score that ChatGPT/GPT4 assigns to assess how effectively the AI Assistant's response aligns with the user's instructions. \\

        \bottomrule
    \end{tabular}
    }
    \caption{Components of \texttt{Generators} and \texttt{Selectors} modules of \raisebox{-1.8pt}{\includegraphics[scale=0.07]{figures/EasyInstruct.png}}. The instruction generation methods implemented in \texttt{Generators} are categorized into three groups, based on their respective seed data sources: chat data, corpus, and knowledge graphs. The evaluation metrics in \texttt{Selecors} are divided into four categories, based on the principle of their implementation: statistics-based, n-gram-based, structure-based, and LM-based.}
    \vspace{-0.1in}
    \label{tab:modules}
\end{table*}

\subsection{Generators}

The \texttt{Generators} module streamlines the process of instruction generation, enabling automated generation of instruction data based on seed data, where seed data can be sourced from either chat data, corpus, or Knowledge Graphs.
As listed in Table~\ref{tab:modules}, the instruction generation methods implemented in \texttt{Generators} are categorized into three groups, based on their respective seed data sources.

\vspace{-0.05in}
\paragraph{Chat Data.}
Early work~\citep{Self-Instruct} randomly samples a few instructions from a human-annotated seed tasks pool as demonstrations and then, prompts an LLM to generate more instructions and corresponding input-output pairs. 
Due to its adaptability, \textit{Self-Instruct} remains the prevailing preference among automated instruction generation methods.
Similarly, starting with an initial set of instructions, \textit{Evol-Instruct}~\citep{WizardLM} incrementally upgrades them into more complex instructions by prompting an LLM with specific prompts. 
In contrast to the \textit{Self-Instruct} generation approach, \textit{Evol-Instruct} allows for the adjustment of the difficulty and intricacy of the instructions it produces.

\vspace{-0.05in}
\paragraph{Corpus.}
Given an unannotated corpus, \textit{Instruction Backtranslation}~\citep{Backtranslation} creates an instruction following training instance by predicting an instruction that would be correctly answered by a paragraph in the document or corpus.
Considering the mixed quality of human-written web text and the presence of noise in generated content, only the highest quality instances are reserved.

\vspace{-0.05in}
\paragraph{Knowledge Graphs.}
Incorporating existing knowledge graphs, \textit{KG2Instruct}~\citep{InstructIE} generates Information Extraction (IE) instruction datasets. 
To enhance the generalizability of instructions, a random sampling approach is utilized based on human-crafted instruction templates.

\begin{figure}[htbp]
\centering
\begin{minipage}{\linewidth}
\usemintedstyle{vs}
\definecolor{bg}{rgb}{0.96,0.96,0.96}
\begin{minted}[breaklines, firstnumber=last, bgcolor=bg,
fontfamily=courier,fontsize=\scriptsize]{python}
from easyinstruct import SelfInstructGenerator
from easyinstruct import GPTScoreSelector
from easyinstruct.utils.api import set_openai_key

# Step1: Set your own API-KEY
set_openai_key("YOUR-KEY")

# Step2: Declare a generator class
generator = SelfInstructGenerator(
    data_format = "alpaca", 
    seed_tasks_path = "seed_tasks.jsonl", 
    generated_instances_path = "generation.jsonl",
    num_instructions_to_generate=100, 
    engine = "gpt-3.5-turbo",
)

# Step3: Generate self-instruct data
generator.generate()

# Step4: Declare a selector class
selector = GPTScoreSelector(
    source_file_path = "generation.jsonl", 
    engine = "gpt-3.5-turbo", 
    threshold = 4,
)

# Step5: Process raw data
selector.process()
\end{minted}
\end{minipage}
\caption{
A running example of instruction generation and selection in \raisebox{-1.8pt}{\includegraphics[scale=0.07]{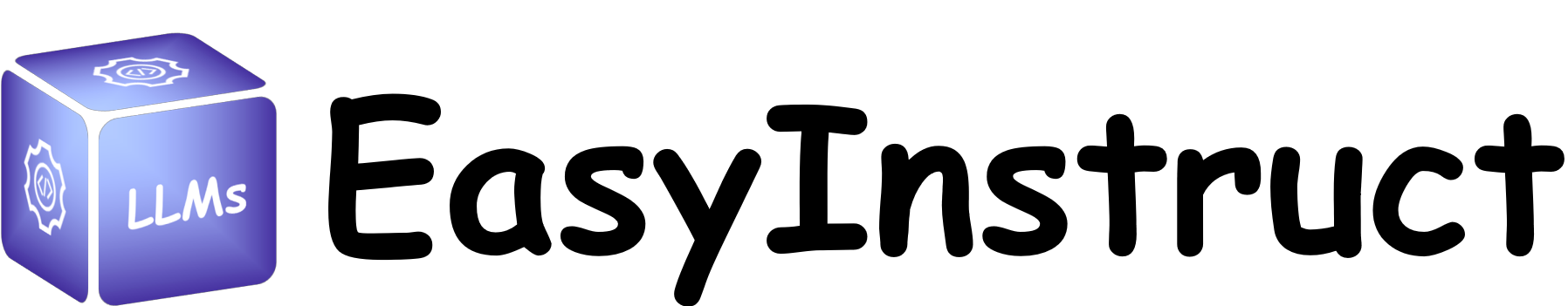}}.}
\vspace{-0.1in}
\label{fig:code_example}
\end{figure}

EasyInstruct has implemented the existing methods above to facilitate future research and systematic comparison of automated generation of instruction data.
Furthermore, the flexibility of the \texttt{Generators} module allows practitioners to select the appropriate generator and make further modification that best suits their specific needs. 
A running example of using a \texttt{Generator} class in EasyInstruct is shown in Figure~\ref{fig:code_example}.

\subsection{Selectors}

The \texttt{Selectors} module is designed to streamline the process of filtering instructions, enabling the curation of instruction datasets from raw instruction data.
This raw data might originate from publicly accessible instruction datasets or be synthesised in advence by the \texttt{Generators} module.
Table~\ref{tab:modules} provides a comprehensive overview of various metrics for instruction quality evaluation.
We divide the evaluation metrics into four categories based on the principle of their implementation: statistics-based, n-gram-based, structure-based and LM-based.
All \texttt{Selector} classes derive from a common base class, \texttt{BaseSelector}.
It includes fundamental attributes and abstract methods such as loading, processing, and dumping of data.
In EasyInstruct, multiple \texttt{Selectors} can be grouped for convenient usage, which allows users to achieve more concise and readable code.
A running example of using a \texttt{Selector} class is shown in Figure~\ref{fig:code_example}.

\subsection{Prompts}

The \texttt{Prompts} module standardizes the instruction prompting step, in which user requests are constructed as instruction prompts and sent to specific LLMs to obtain responses.
Utilizing the \texttt{Prompts} module with a series of well-designed and refined prompts enhances the ability of \texttt{Generators} and \texttt{Selectors} to effectively fulfill their respective functions.
Similar to \texttt{Selectors}, all \texttt{Prompts} classes inherit from a common base class, \texttt{BasePrompt}, which
includes necessary attributes and abstract methods.
In the mentioned base class, there are functionalities provided for building prompts, requesting generation results from LLMs, and parsing the responses received from LLMs.
The base class also provides mechanisms to handle error conditions and exceptions that may occur during the whole process.
Users can inherit from the base class and customize or extend its functionality based on their specific requirements.
We also equip EasyInstruct with various prompting techniques and application adaptions (e.g. Chain-of-Thought, Information Extraction, Multimodal, etc.) by providing a consistent and standardized interface, enabling efficient instruction prompting for LLMs.

\section{Evaluation}

In terms of evaluation, we will introduce the experiment setups and illustrate the empirical results of multiple modules implemented in EasyInstruct to demonstrate its capability.

\begin{table*}[htbp]
    \renewcommand\arraystretch{1.2}
    \centering
    \resizebox{\linewidth}{!}{
        \begin{tabular}{p{18cm}}
        \toprule
        \colorbox{light-gray}{\textbf{[Instruction]}} Describe the common theme of the following three animals. \\
        \colorbox{light-gray}{\textbf{[Output]}} The common theme among the penguin, polar bear and walrus is their ability to survive in cold climates. All three animals have adapted to thrive in cold Arctic regions, with a heavy layer of blubber or thick fur to keep them warm and plenty of blubber to provide sustenance. \\
        \colorbox{light-gray}{\textbf{[Scores]}} \{\textcolor{teal}{"mtld\_score": 22.68}, \textcolor{orange}{"avg\_rouge\_score": 0.16}, \textcolor{violet}{"ppl\_score": 22.57}, \textcolor{brown}{ "gpt\_score": 4}\} \\
        \midrule
        \colorbox{light-gray}{\textbf{[Instruction]}} Pretend that you are the subject of the following poem. Use the first person voice to write a response to the poem. \\
        \colorbox{light-gray}{\textbf{[Input]}} The gentle swell of the sea,\char92 n Cradling the shore in its serenity. \\
        \colorbox{light-gray}{\textbf{[Output]}} I can hear the sound of the waves crashing against the shore. The sea's serenity is like a soothing cradle that brings me peace and tranquility. \\
        \colorbox{light-gray}{\textbf{[Scores]}} \{\textcolor{teal}{"mtld\_score": 27.10}, \textcolor{orange}{"avg\_rouge\_score": 0.16}, \textcolor{violet}{"ppl\_score": 26.02}, \textcolor{brown}{"gpt\_score": 4}\} \\
        \midrule
        \colorbox{light-gray}{\textbf{[Instruction]}} Create a function to calculate the total cost of a book order, given the unit price, quantity, shipping cost, and tax rate. \\
        \colorbox{light-gray}{\textbf{[Input]}} Unit Price: 10.99\char92 n Quantity: 3\char92 n Shipping Cost: 5\char92 n Tax Rate: 0.03 \\
        \colorbox{light-gray}{\textbf{[Output]}} def calculateCost(unitPrice, quantity, shippingCost, taxRate):\par
        \qquad\qquad\qquad totalPrice = (unitPrice * quantity) + shippingCost \par
        \qquad\qquad\qquad totalTax = totalPrice * taxRate \par
        \qquad\qquad\qquad totalCost = totalPrice + totalTax \par
        \qquad\qquad\qquad return totalCost \\
        \colorbox{light-gray}{\textbf{[Scores]}} \{\textcolor{teal}{"mtld\_score": 67.76}, \textcolor{orange}{"avg\_rouge\_score": 0.17}, \textcolor{violet}{"ppl\_score": 18.95}, \textcolor{brown}{"gpt\_score": 5}\} \\
        \bottomrule
        \end{tabular}
    }
    \caption{Examples of the selected instructions in \textbf{\textit{easyinstruct\_5k}}, and their corresponding evaluation scores given by the \texttt{Selectors} module.}
    \vspace{-0.1in}
    \label{tab:case}
\end{table*}

\subsection{Experiment Setups}

\paragraph{Instruction Datasets.}
We adopt the popular \textit{Self-Instruct}~\citep{Self-Instruct} and \textit{Evol-Instruct}~\citep{WizardLM} methods implemented in EasyInstruct to synthesize instruction datasets, containing instructions paired with instance inputs and outputs separately.
We mainly consider four instruction datasets as follows: (a) \textbf{\textit{self\_instruct\_5k}} is constructed by employing the \textit{Self-Instruct} method to distill instruction data from text-davinci-003; (b) \textbf{\textit{alpaca\_data\_5k}} is randomly sampled from the Alpaca dataset~\citep{Alpaca}; (c) \textbf{\textit{evol\_instruct\_5k}} is constructed by employing the \textit{Evol-Instruct} method; (d) \textbf{\textit{easyinstruct\_5k}} is collected by integrating the three instruction datasets above and applying multiple \texttt{Selectors} in EasyInstruct to extract high-quality instruction datasets.
We detail the hyper-parameters for selection in a configuration file, as shown in Appendix~\ref{app:shell_script}.

\paragraph{Implementation Setups.}
To conduct the experiments on the effect of instruction datasets, we adopt a LLaMA2 (7B) model~\citep{Llama2}, a popular open-sourced LLM backbone widely used for instruction-tuning.
We fine-tune the LLMs with LoRA~\citep{LoRA} in the format following Alpaca~\citep{Alpaca} with KnowLM\footnote{\url{https://github.com/zjunlp/KnowLM}}.
All hyper-parameters for training can be found in Appendix \ref{app:hp}.
The evaluation is conducted by comparing the generated results from different fine-tuned models based on the AlpacaFarm~\citep{AlpacaFarm} evaluation set.
Following AlpacaFarm, for each comparison, we employ ChatGPT as the evaluator to automatically compare two outputs from different models and label which one they prefer, reporting the win rate as the evaluation metric.
For both instruction tuning and evaluation, we adopt the same prompt templates used by Alpaca-LoRA\footnote{\url{https://github.com/tloen/alpaca-lora}}, as shown in Appendix~\ref{app:prompt_template}.

\subsection{Experiment Results}

\begin{figure}[htbp]
    \centering
    \resizebox{\linewidth}{!}{
        \includegraphics{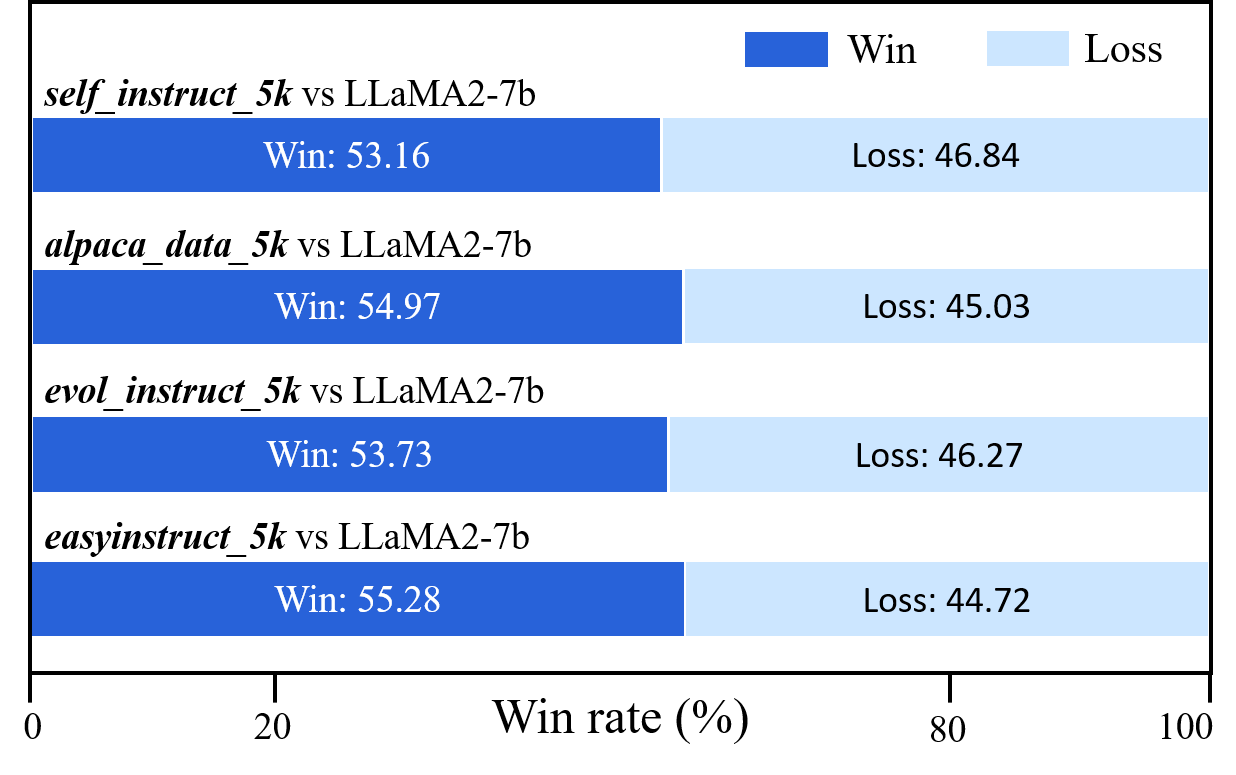}
    }
    \caption{Results of models fine-tuned on four distinct instruction datasets against those from the base LLaMA2 (7B) model, using the AlpacaFarm evaluation set for assessment.}
    \vspace{-0.1in}
    \label{fig:base_results}
\end{figure}

\paragraph{Main Results.}
We compare the generated outputs from models fine-tuned separately on the four instruction datasets with the outputs from the base version of the LLaMA2 (7B) model on the AlpacaFarm evaluation set.
As depicted in Figure~\ref{fig:base_results}, there are improvements in the win rate metric for all the settings.
Moreover, the model performs optimally under the \textbf{\textit{easyinstruct\_5k}} setting, indicating the importance of a rich instruction selection strategy.

\begin{figure}[htbp]
    \centering
    \resizebox{0.9\linewidth}{!}{
        \includegraphics{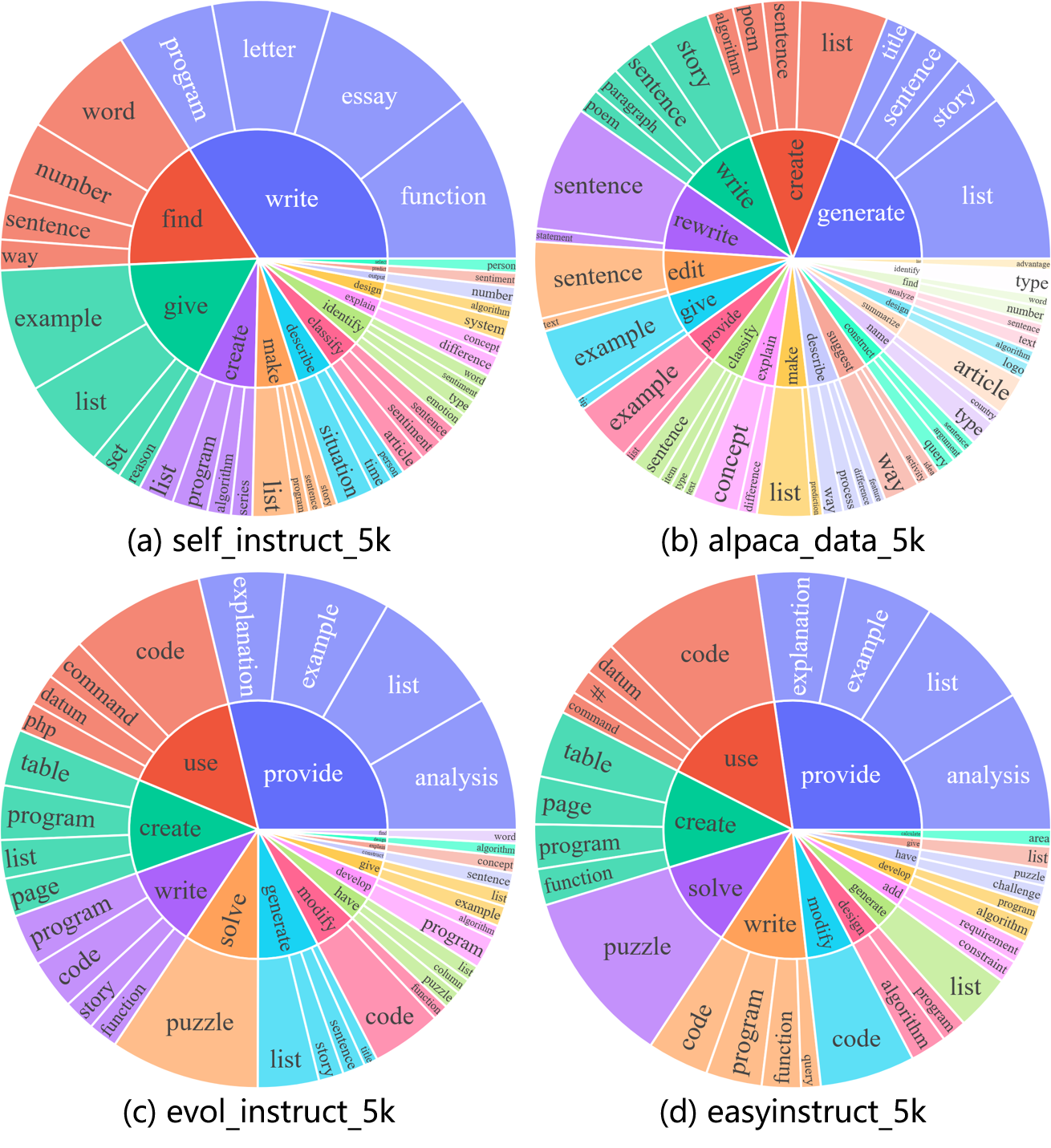}
    }
    \caption{(Inner circle refers to the top 20 most prevalent root verbs and outer circle indicates their top 4 direct nouns in the generated instruction datasets considered in the experiments.}
    \vspace{-0.1in}
    \label{fig:diversity}
\end{figure}

\paragraph{Instruction Diversity.}
To study the diversity of the instruction datasets considered in our experiments, we identify the verb-noun structure in the generated instructions and plot the top 20 most prevalent root verbs and their top 4 direct nouns in Figure~\ref{fig:diversity}, following the approach of \citet{Self-Instruct}.
Overall, we see a wide range of intents and textual formats within these instructions.

\paragraph{Case Study.}
To conduct a qualitative evaluation of EasyInstruct, we sample several instruction examples selected by the \texttt{Selectors} module in \textbf{\textit{easyinstruct\_5k}} for the case study.
We also attach the corresponding evaluation scores for each of these instruction examples, as shown in Table~\ref{tab:case}.
We observe that the selected instructions often possess fluent language and meticulous logic.

\section{Conclusion and Future Work}

We present \ours, an easy-to-use instruction processing framework for LLMs.
EasyInstruct can combine chat data, corpus, KGs and LLMs as an automated instruction generation tool, reducing the cost of manual data annotation.
Additionally, EasyInstruct integrates a diverse set of instruction selection tools to optimize the diversity and distribution of instruction data, thereby improving the quality of fine-tuning data.
EasyInstruct is designed to be easy to extend, and we will continue to update new features (e.g., knowledgeable synthetic data generation) to keep pace with the latest research.
We expect EasyInstruct to be a helpful framework for researchers and practitioners to facilitate their work of instruction tuning on LLMs.

\section*{Limitations}

In this paper, we are committed to unifying all phases of instruction data processing including instruction generation, selection, and prompting.
Despite our efforts, this paper may still have some remaining limitations.

\paragraph{The Scope of Instruction Selection Methods.}
We implement various instruction selection methods within the \texttt{Selectors} module.
Based on the evaluation metrics utilized and the model base employed, the implemented instruction data selection methods can be divided into three categories: methods based on a system of indicators, methods utilizing powerful LLMs like ChatGPT, and methods employing small models~\citep{selection-survey}.
However, another line of work~\citep{IFD,nuggets,Backtranslation,diverseeval,tegit,active_it} employs trainable LLMs like LLaMA for computation formulas in instruction selection processes, which are not integrated into the \texttt{Selectors} module.
Although our design choice is to decouple instruction processing and model training into two separate phases, we regard it as a limitation that may be addressed by future work.

\paragraph{Statistics for evaluating efficiency.}
In our evaluation, we fine-tune a LLaMA2 (7B) model utilizing multiple modules implemented in EasyInstruct.
Compared to models fine-tuned on other instruction datasets constructed without EasyInstruct, our model achieves optimal results, demonstrating EasyInstruct's capability.
Although we also qualitatively demonstrate the ease of writing code for instruction processing with multiple code samples and configuration files using EasyInstruct, a limitation is the lack of appropriate statistics for quantitatively evaluating efficiency.

\section*{Acknowledgments}

We would like to express gratitude to the anonymous reviewers for their kind comments. 
This work was supported by the National Natural Science Foundation of China (No. 62206246, No. NSFCU23B2055, No. NSFCU19B2027), the Fundamental Research Funds for the Central Universities (226-2023-00138), Zhejiang Provincial Natural Science Foundation of China (No. LGG22F030011), Yongjiang Talent Introduction Programme (2021A-156-G), CCF-Baidu Open Fund, Information Technology Center and State Key Lab of CAD\&CG, Zhejiang University.

\bibliography{custom}
\bibliographystyle{acl_natbib}

\newpage
\appendix

\section{Appendix}
\label{sec:appendix}

\subsection{Installation}

Currently, EasyInstruct offers three installation options, each accompanied by its corresponding installation script.
Users can choose the option that best suits their specific requirements.

\subsubsection{Installation from GitHub Repository}

The first option is to install the latest version of EasyInstruct from the GitHub repository. The installation script is shown in Figure~\ref{fig:installation_github}.

\begin{figure*}[htbp]
\centering
\begin{minipage}{\linewidth}
\usemintedstyle{emacs}
\definecolor{bg}{rgb}{0.96,0.96,0.96}
\begin{minted}[breaklines, firstnumber=last, bgcolor=bg, fontfamily=courier,fontsize=\footnotesize]{bash}
pip install git+https://github.com/zjunlp/EasyInstruct@main
\end{minted}
\end{minipage}
\caption{Installation script from Github repository.}
\label{fig:installation_github}
\end{figure*}

\subsubsection{Installation for Local Development}

The second option is to download the source code for local development. The installation script is shown in Figure~\ref{fig:installation_develop}.

\begin{figure}[htbp]
\centering
\begin{minipage}{\linewidth}
\usemintedstyle{emacs}
\definecolor{bg}{rgb}{0.96,0.96,0.96}
\begin{minted}[breaklines, firstnumber=last, bgcolor=bg, fontfamily=courier,fontsize=\footnotesize]{bash}
git clone https://github.com/zjunlp/EasyInstruct
cd EasyInstruct
pip install -e .
\end{minted}
\end{minipage}
\caption{Installation script for local development.}
\vspace{-0.15in}
\label{fig:installation_develop}
\end{figure}

\subsubsection{Installation from PyPI}

The third option is to install the package from The Python Package Index (PyPI), which may not be the latest version but still supports most of the features. The installation script is shown in Figure~\ref{fig:installation_pip}.

\begin{figure}[htbp]
\centering
\begin{minipage}{\linewidth}
\usemintedstyle{emacs}
\definecolor{bg}{rgb}{0.96,0.96,0.96}
\begin{minted}[breaklines, firstnumber=last, bgcolor=bg, fontfamily=courier,fontsize=\footnotesize]{bash}
pip install easyinstruct
\end{minted}
\end{minipage}
\caption{Installation script using PyPI.}
\vspace{-0.15in}
\label{fig:installation_pip}
\end{figure}

\subsection{Quick-start}

We provide two ways for users to quickly get started with EasyInstruct. Users can either use the shell script or the Gradio app based on their specific needs.

\subsubsection{Shell Script}
\label{app:shell_script}

\paragraph{Step1: Prepare a configuration file.}
Users can easily configure the parameters of EasyInstruct in a YAML-style file or just quickly use the default parameters in the configuration files we provide. 
Figure~\ref{fig:self_instruct_configuration} is an example of the configuration file for Self-Instruct.

\begin{figure}[htbp]
\centering
\begin{minipage}{\linewidth}
\usemintedstyle{emacs}
\definecolor{bg}{rgb}{0.96,0.96,0.96}
\begin{minted}[breaklines, firstnumber=last, bgcolor=bg, fontfamily=courier,fontsize=\footnotesize]{yaml}
generator:
  SelfInstructGenerator:
    target_dir: data/generations/
    data_format: alpaca
    seed_tasks_path: data/seed_tasks.jsonl
    generated_instructions_path: generated_instructions.jsonl
    generated_instances_path: generated_instances.jsonl
    num_instructions_to_generate: 100
    engine: gpt-3.5-turbo
    num_prompt_instructions: 8
\end{minted}
\end{minipage}
\caption{Example configuration file of \texttt{Generators}.}
\vspace{-0.15in}
\label{fig:self_instruct_configuration}
\end{figure}

\begin{figure}[htbp]
\centering
\begin{minipage}{\linewidth}
\usemintedstyle{emacs}
\definecolor{bg}{rgb}{0.96,0.96,0.96}
\begin{minted}[breaklines, firstnumber=last, bgcolor=bg, fontfamily=courier,fontsize=\footnotesize]{yaml}
selector:
  source_file_path: 
  target_dir: data/selections/
  target_file_name: case.jsonl
  LengthSelector:
    min_instruction_length: 3
    max_instruction_length: 150
    min_response_length: 1
    max_response_length: 350
  Deduplicator:
  RougeSelector:
    threshold: 0.7
  GPTScoreSelector:
    engine: gpt-3.5-turbo
    threshold: 4
  MTLDSelector:
    ttr_threshold: 0.72
    min_mtld: 8
    max_mtld: 22
  PPLSelector:
    threshold: 200
    model_name: gpt2
    device: cuda
  RandomSelector:
    num_instructions_to_sample: 100
    seed: 42
\end{minted}
\end{minipage}
\caption{Example configuration file of \texttt{Selectors}.}
\vspace{-0.15in}
\label{fig:selectors_configuration}
\end{figure}

\paragraph{Step2: Run the shell script.}
Users should first specify the configuration file and provide their own OpenAI API key. Then, run the following shell script in Figure~\ref{fig:shell_script} to launch the instruction generation or selection process.

\begin{figure}[htbp]
\centering
\begin{minipage}{\linewidth}
\usemintedstyle{emacs}
\definecolor{bg}{rgb}{0.96,0.96,0.96}
\begin{minted}[breaklines, firstnumber=last, bgcolor=bg, fontfamily=courier,fontsize=\footnotesize]{bash}
config_file=""
openai_api_key=""

python demo/run.py \
    --config  $config_file\
    --openai_api_key $openai_api_key \
\end{minted}
\end{minipage}
\caption{Shell script for quick-start of EasyInstruct.}
\vspace{-0.15in}
\label{fig:shell_script}
\end{figure}

\subsubsection{Gradio App}

We provide a Gradio app for users to quickly get started with EasyInstruct. 
Users can choose to launch the Gradio App locally on their own machines or alternatively, they can also try the hosted Gradio App\footnote{\url{https://huggingface.co/spaces/zjunlp/EasyInstruct}.} that we provide on HuggingFace Spaces.






\subsection{Detailed Hyper-Parameters}
\label{app:hp}

See Table~\ref{tab:hp}.

\begin{table}[htbp]
    \centering
    \begin{tabular}{cc}
    \toprule
    \textbf{Name} &  \textbf{LLaMA-2-7b}  \\
    \toprule
    batch\_size & 256 \\
    micro\_batch\_size & 8 \\
    epochs & 3 \\
    learning rate & 3e-4 \\
    cutoff\_len & 512 \\
    val\_set\_size & 1,000 \\
    lora\_r & 16 \\
    lora\_alpha & 32 \\
    lora\_dropout & 0.05 \\
    \bottomrule
    \end{tabular}
    \caption{Detailed hyper-parameters we use in experiments.}
    \label{tab:hp}
\end{table}




\subsection{Prompt Template for Instruction Tuning}
\label{app:prompt_template}

For both training and evaluation, we utilize the same prompt templates used by Alpaca-LoRA, shown in Table~\ref{tab:prompt_instruction_tuning}.

\begin{table}[htbp]
    \centering
    \resizebox{\linewidth}{!}{
    \begin{tabular}{p{8cm}}
    \toprule
    \textbf{Prompt Template for Instruction Tuning}\\
    \midrule
    \rowcolor{light-gray}
    \textbf{\textit{Prompt with Input:}} \\
    Below is an instruction that describes a task, paired with an input that provides further context. Write a response that appropriately completes the request.\par \\
    \#\#\# Instruction:\par 
    \textbf{\{instruction\}}\par \\
    \#\#\# Input:\par
    \textbf{\{input\}}\par \\
    \#\#\# Response: \\
    \midrule
    \rowcolor{light-gray}
    \textbf{\textit{Prompt without Input:}} \\
    Below is an instruction that describes a task. Write a response that appropriately completes the request.\par \\
    \#\#\# Instruction:\par
    \textbf{\{instruction\}}\par \\
    \#\#\# Response: \\
    \bottomrule
    \end{tabular}
    }
    \caption{Prompt Template for instruction tuning.}
    \vspace{-0.15in}
    \label{tab:prompt_instruction_tuning}
\end{table}

\subsection{API Services Available in EasyInstruct}
\label{app:api}

Table~\ref{tab:provider} lists a range of API service providers and their corresponding LLM products that are currently available in EasyInstruct.

\begin{table*}[htbp]
\renewcommand\arraystretch{1.10}
\centering
\resizebox{\linewidth}{!}{
\begin{tabular}{lp{11.5cm}c}
\toprule
\textbf{Model} & \textbf{Description} & \textbf{Default Version} \\ \midrule
\rowcolor{light-gray}
\multicolumn{3}{l}{\textbf{\textit{OpenAI}}} \\
\multirow{2}{*}{GPT-3.5}         & A set of models that improve on GPT-3 and can understand as well as generate natural language or code.   & \multirow{2}{*}{\texttt{gpt-3.5-turbo}}          \\ 
\multirow{2}{*}{GPT-4}           & A set of models that improve on GPT-3.5 and can understand as well as generate natural language or code. & \multirow{2}{*}{\texttt{gpt-4}}                  \\ \midrule
\rowcolor{light-gray}
\multicolumn{3}{l}{\textbf{\textit{Anthropic}}} \\ 
\multirow{2}{*}{Claude}        & A next-generation AI assistant based on Anthropic’s research into training helpful, honest, and harmless AI systems. & \multirow{2}{*}{\texttt{claude-2}}                  \\ 
Claude-Instant  &  A lighter, less expensive, and much faster option than Claude.  &  \texttt{claude-instant-1}               \\ \midrule
\rowcolor{light-gray}
\multicolumn{3}{l}{\textbf{\textit{Cohere}}} \\
 \multirow{3}{*}{Command}         & An instruction-following conversational model that performs language tasks with high quality, more reliably, and with a longer context than cohere's base generative models.  &   \multirow{3}{*}{\texttt{command}}               \\ 
Command-Light   & A smaller, faster version of Command. Almost as capable, but a lot faster. &  \texttt{command-light}                  \\ \bottomrule
\end{tabular}
}
\caption{API service providers and their corresponding LLM products that are currently available in \raisebox{-1.8pt}{\includegraphics[scale=0.07]{figures/EasyInstruct.png}}.}
\vspace{-0.15in}
\label{tab:provider}
\end{table*}

\subsection{Example features in the \texttt{Prompts} module}
\label{app:prompts_example}

\begin{figure*}[htbp]
    \centering
    \resizebox{\linewidth}{!}{
        \includegraphics{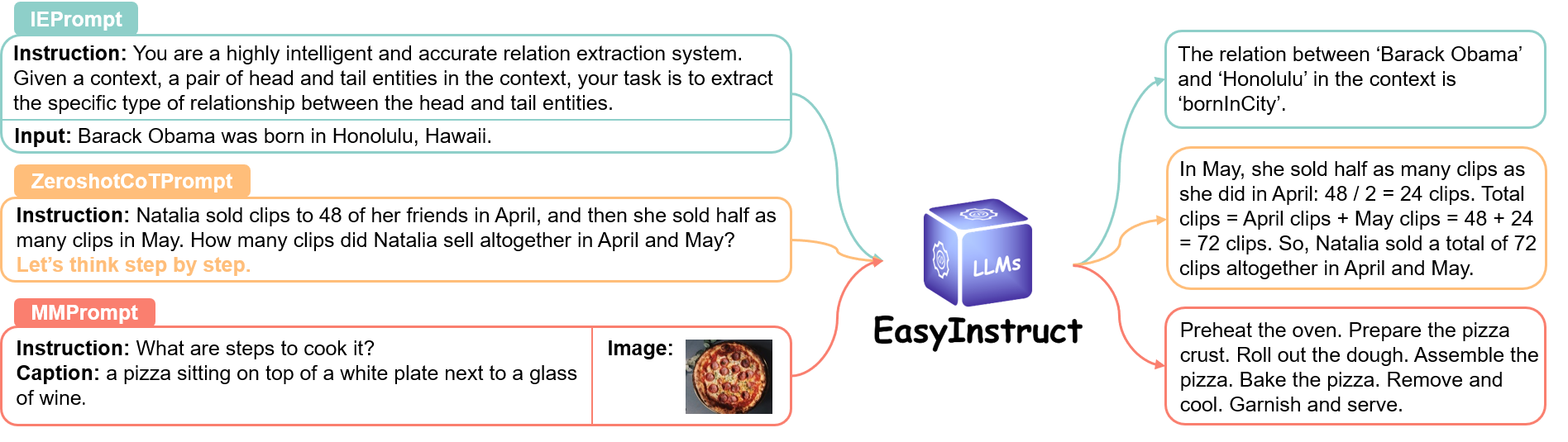}
    }
    \caption{Example features in the \texttt{Prompts} module, including Information Extraction, Chain-of-Thought Reasoning, and Multimodal Prompting.}
    \vspace{-0.15in}
    \label{fig:prompt}
\end{figure*}

\subsection{Acknowledgements}
We thank the developers of the self-instruct\footnote{\url{https://github.com/yizhongw/self-instruct}} library for their significant contributions to the NLP community.
We thank the LLaMA team for providing us access to the models, and open-source projects, including Alpaca\footnote{\url{https://github.com/tatsu-lab/stanford_alpaca}}, Alpaca-LoRA\footnote{\url{https://github.com/tloen/alpaca-lora}} and AlpacaEval\footnote{\url{https://github.com/tatsu-lab/alpaca_eval}}.

\end{document}